\documentclass[conference]{IEEEtran}
\IEEEoverridecommandlockouts
\usepackage{cite}
\usepackage{amsmath,amssymb,amsfonts}
\usepackage{booktabs}
\usepackage{multirow}
\usepackage{algorithmic}
\usepackage{graphicx}
\usepackage{textcomp}
\usepackage{xcolor}
\def\BibTeX{{\rm B\kern-.05em{\sc i\kern-.025em b}\kern-.08em
    T\kern-.1667em\lower.7ex\hbox{E}\kern-.125emX}}
\begin{document}

\title{WiFi2Cap: Semantic Action Captioning from Wi-Fi CSI via Limb-Level Semantic Alignment}

\author{\IEEEauthorblockN{Tzu-Ti Wei, Chu-Yu Huang, Yu-Chee Tseng, and Jen-Jee Chen}
\IEEEauthorblockA{\textit{PAIRLab, College of AI, National Yang Ming Chiao Tung University, Taiwan, R.O.C}\\
\{a2699560.ai09, apple32112311.ai12, yctseng, jenjee\}@nycu.edu.tw} \\
}

\maketitle

\begin{abstract}
Privacy-preserving semantic understanding of human activities is important for indoor sensing, yet existing Wi-Fi CSI-based systems mainly focus on pose estimation or predefined action classification rather than fine-grained language generation. Mapping CSI to natural-language descriptions remains challenging because of the semantic gap between wireless signals and language and direction-sensitive ambiguities such as left/right limb confusion. We propose \emph{WiFi2Cap}, a three-stage framework for generating action captions directly from Wi-Fi CSI. A vision--language teacher learns transferable supervision from synchronized video--text pairs, and a CSI student is aligned to the teacher's visual space and text embeddings. To improve direction-sensitive captioning, we introduce a Mirror-Consistency Loss that reduces mirrored-action and left--right ambiguities during cross-modal alignment. A prefix-tuned language model then generates action descriptions from CSI embeddings. We also introduce the \emph{WiFi2Cap Dataset}, a synchronized CSI--RGB--sentence benchmark for semantic captioning from Wi-Fi signals. Experimental results show that WiFi2Cap consistently outperforms baseline methods on BLEU-4, METEOR, ROUGE-L, CIDEr, and SPICE, demonstrating effective privacy-friendly semantic sensing.
\end{abstract}

\section{Introduction}

Generating natural-language descriptions of human activities is a key capability for human-centered sensing and interaction. Recent vision--language models and contrastive pretraining (e.g., CLIP~\cite{radford2021learning}) have substantially improved captioning from visual inputs, but deploying cameras in private indoor spaces (e.g., bedrooms and hospital wards) raises serious privacy concerns.

This motivates privacy-preserving alternatives that avoid recording identifiable appearance information. Prior work has explored non-visual modalities such as mmWave radar, infrared sensing, and Wi-Fi signals~\cite{li2024mmskeleton,sengupta2020mm,ryoo2017privacy,gade2014thermal,chiu2023privacy,ren2022gopose,yang2024privacy,yan2024personwifi3d}. Among them, Wi-Fi Channel State Information (CSI) is robust to occlusion and lighting, and has enabled fine-grained perception such as pose estimation~\cite{wang2019personwifi,yan2024personwifi3d,zhao2018rfpose}.

However, existing CSI-based systems largely target low- or mid-level outputs, such as joint coordinates, segmentation masks, or predefined action labels, rather than free-form and fine-grained language generation. As a result, mapping numerical radio signals to natural-language descriptions remains challenging for two reasons. First, there is a substantial semantic gap between CSI and language, together with limited paired CSI--text supervision. Second, action captioning requires preserving direction-sensitive semantics, such as left/right limbs and mirrored movements, which are easily confused during cross-modal alignment.

\begin{figure*}[htbp]
    \centering
    \includegraphics[width=0.9\textwidth]{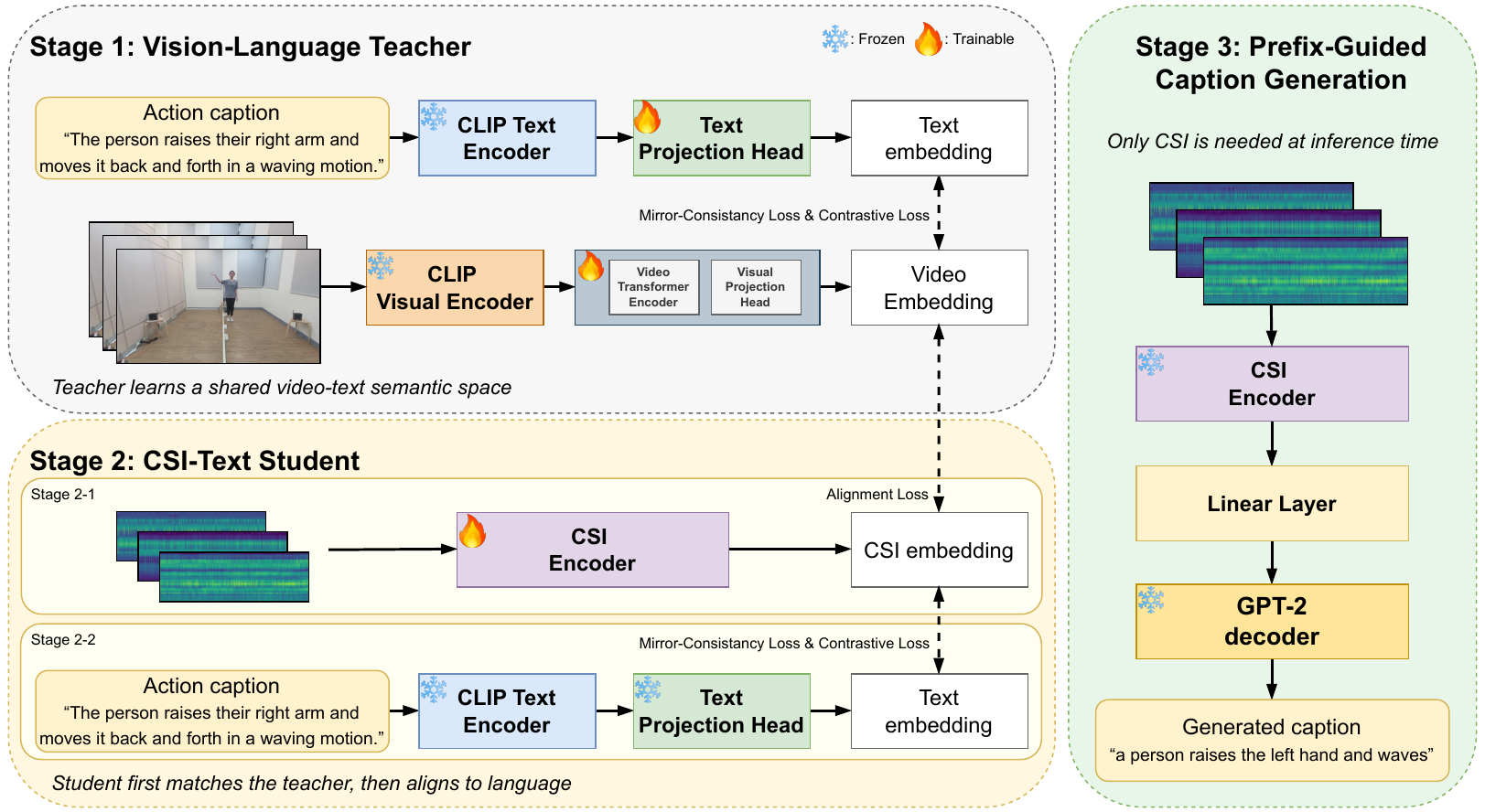}
    \caption{\textbf{WiFi2Cap framework.}
    (a) Stage 1: Vision--language teacher trained with contrastive learning and Mirror-Consistency.
    (b) Stage 2: CSI--text student trained via teacher-guided visual alignment and CSI--text contrastive learning with Mirror-Consistency.
    (c) Stage 3: Prefix-guided language generation.
    }
    \label{fig:framework}
\end{figure*}

To address these challenges, we study the largely unexplored problem of \emph{semantic action captioning} directly from Wi-Fi CSI. We propose \emph{WiFi2Cap}, a three-stage framework in which a data-rich vision--language teacher transfers semantic knowledge to a CSI student through contrastive alignment, followed by caption generation with a frozen autoregressive language model via prefix tuning (Fig.~\ref{fig:framework}). To reduce frequent confusions between mirrored actions, such as left--right limb ambiguity, we further introduce a \emph{Mirror-Consistency Loss} during alignment.

To support both training and evaluation, we also introduce the \emph{WiFi2Cap Dataset}, a synchronized benchmark for CSI-based semantic captioning that combines Wi-Fi CSI, RGB videos, and sentence-level action descriptions. The dataset covers 100 action classes, with each instance recorded as a 5-second clip and captured with one transmitter and three receivers.

\noindent\textbf{Contributions.} This paper makes three contributions:
\begin{itemize}
\item \textbf{Wi-Fi CSI action captioning:} We propose \emph{WiFi2Cap}, a framework that generates fine-grained action captions directly from CSI by combining visual knowledge transfer with contrastive learning.
\item \textbf{Mirror-Consistency for direction-sensitive semantics:} We identify left/right limb confusion and mirrored-action ambiguity as a critical failure mode in cross-modal action captioning, and introduce a mirror-consistency objective to explicitly enforce limb-aware, direction-sensitive alignment.
\item \textbf{Dataset and evaluation:} We introduce the \emph{WiFi2Cap Dataset}, a synchronized CSI--RGB--sentence benchmark tailored to semantic caption generation, and demonstrate consistent gains over CSI-only baselines on standard captioning metrics.
\end{itemize}
\section{Related Work}

\subsection{RF-Based Human Sensing}
RF sensing has gained increasing attention for device-free human understanding due to its robustness to lighting and occlusion and its privacy advantage over cameras~\cite{nguyen2024noncontact,biase2022markerless}. Beyond Wi-Fi, other privacy-preserving modalities such as mmWave radar and infrared sensing have also been explored for human-centric semantic understanding~\cite{li2024mmskeleton,ryoo2017privacy,chiu2023privacy}, and recent radar--language models further suggest a growing interest in language-grounded semantic interpretation of non-visual signals~\cite{pushkareva2024radar,yuan2026sig2text}. CSI-based systems can recover rich motion cues for tasks such as segmentation and 2D/3D pose estimation~\cite{wang2019personwifi,yan2024personwifi3d,jiang2020wifi3d}, and recent Transformer variants further improve temporal--spectral modeling~\cite{zhou2022csiformer}. However, most RF works still target geometric or categorical outputs (poses or action labels), while free-form \emph{language generation} from Wi-Fi signals remains largely unexplored.

\subsection{Cross-Modal Knowledge Transfer}
Cross-modal transfer learning addresses limited supervision in the target modality by leveraging a stronger source modality~\cite{zhuang2020transfer}. Knowledge distillation transfers soft predictions or intermediate representations from a teacher to a student~\cite{hinton2015distilling}. Recent studies extend this idea to cross-modal representation distillation, where a vision--language teacher defines a semantic embedding space and a student from another modality is trained to align paired samples while separating mismatches~\cite{kim2025cosmos}. WiFi2Cap follows this paradigm by aligning CSI features to a vision--language teacher to bridge the modality gap under scarce CSI--text pairs.

\subsection{Contrastive Learning}
Contrastive objectives such as InfoNCE~\cite{oord2018cpc} and its scalable variants (SimCLR~\cite{chen2020simclr}, MoCo~\cite{he2020moco}) have shown strong transferability, and CLIP~\cite{radford2021learning} demonstrates that large-scale contrastive pretraining can learn highly semantic cross-modal embeddings. A practical challenge for action understanding is \emph{mirror ambiguity} (e.g., left/right limbs), which motivates our mirror-consistency regularization during alignment.

\subsection{Prefix/Prompt Tuning for Language Conditioning}
Prefix/prompt tuning conditions a frozen language model using a small number of trainable parameters, avoiding full fine-tuning. Prefix-Tuning~\cite{li2021prefixtuning} injects learnable key/value prefixes into Transformer attention, while Prompt Tuning~\cite{lester2021power} and P-Tuning~v2~\cite{liu2022p} learn continuous prompts/prefix-like parameters across layers. Following this line, WiFi2Cap maps CSI embeddings into per-layer prefixes via a lightweight MLP for CSI-conditioned caption generation.
\section{WiFi2Cap Dataset}

Existing Wi-Fi sensing datasets mainly target pose estimation or action classification and do not provide the multimodal supervision needed for CSI-based caption generation. To support this task, we construct the \emph{WiFi2Cap} dataset, a synchronized CSI--RGB--sentence benchmark that supports all three stages of our framework.
The dataset covers 100 action categories, and each sample is a 5-second clip with synchronized CSI, RGB observations, and a sentence-level description.

\subsection{Data Collection}
\begin{figure}[t]
 \centering
 \includegraphics[width=0.45\textwidth]{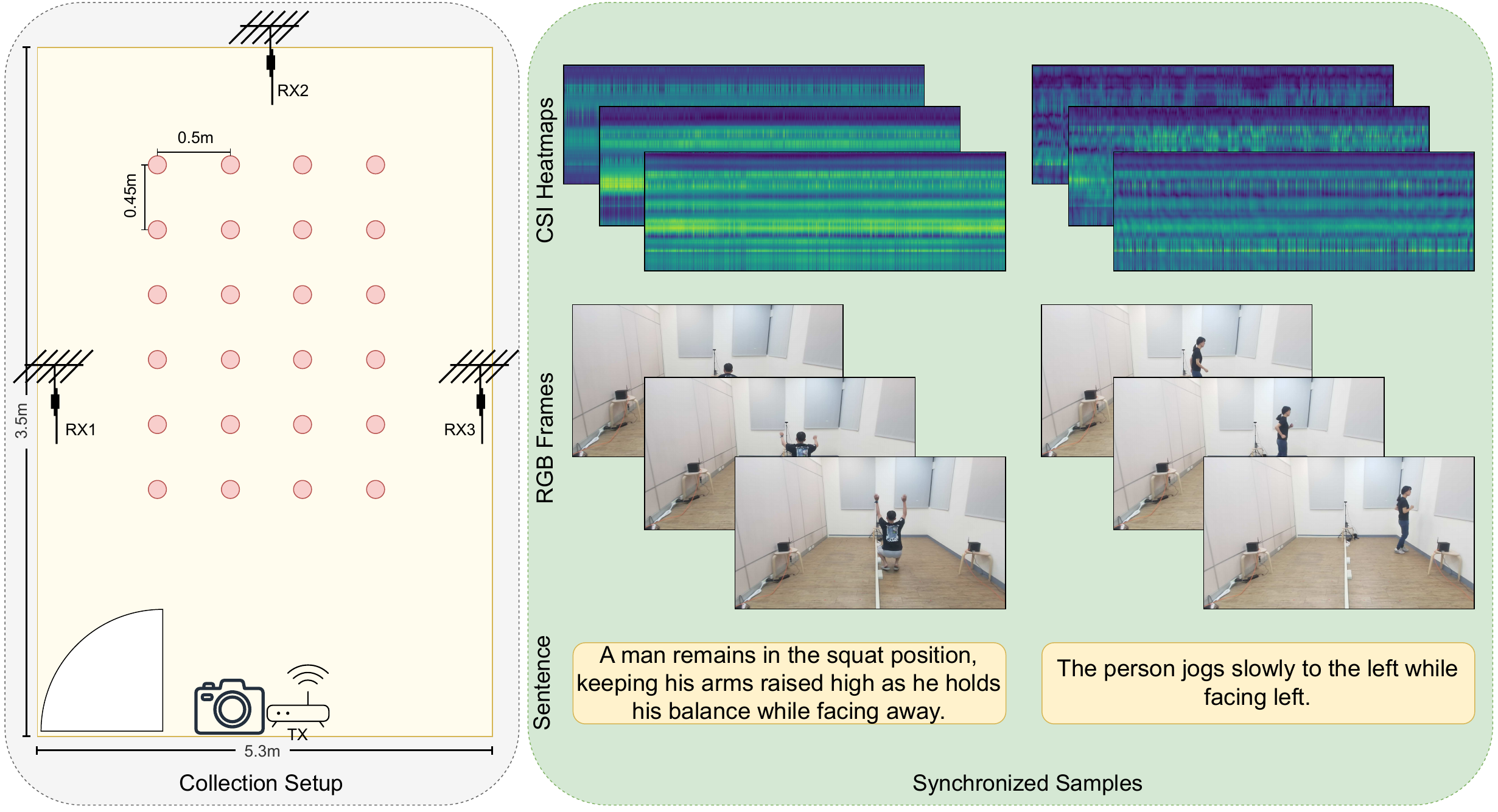}
 \caption{\textbf{Data acquisition setup and multimodal samples.} Left: the physical collection setup, including one transmitter (Tx), three receivers (Rx1--Rx3), the RGB camera, and the $4\times6$ grid of participant positions. Right: synchronized CSI heatmaps, RGB frames, and sentence-level action descriptions.}
 \label{fig:scenario}
\end{figure}

For each action category, participants perform a predefined movement. To introduce controlled spatial diversity, before every recording the participant randomly selects one of 24 standing positions arranged in a $4\times6$ grid. During recording, CSI is collected with one transmitter and three receivers around the participant, while an RGB camera provides the visual reference. All devices remain fixed, and CSI streams are synchronized with RGB videos for aligned training and evaluation (Fig.~\ref{fig:scenario}).

\subsection{Data Annotation}
Each action category is paired with a textual description to supervise both semantic alignment and caption generation. We start from concise action prompts (e.g., ``standing and waving the left hand'') and use a GPT-based large language model (LLM) for wording refinement, while manually verifying all descriptions to avoid semantic drift.

\subsection{CSI Preprocessing}
We decode CSI using PicoScenes~\cite{jiang2020picoscenes}. For each receiver, CSI is represented as complex channel responses over time and decomposed into amplitude and phase sequences. We remove pilot and guard subcarriers and apply standard phase sanitization~\cite{wang2015phasefi}. The three synchronized receivers are treated as multi-view observations, producing the CSI inputs used by the student encoder and caption generator.
\section{WiFi2Cap Framework}

As shown in Fig.~\ref{fig:framework}, WiFi2Cap generates action captions from Wi-Fi CSI in three stages: (i) a \textbf{vision--language teacher} that learns a shared video--text embedding space with contrastive learning and mirror-consistency regularization; (ii) a \textbf{CSI student} that encodes CSI amplitude/phase, fuses them with a gating module, and aligns CSI embeddings to the teacher's visual/text embeddings; and (iii) \textbf{prefix-guided generation} that conditions a frozen language model by mapping CSI embeddings into per-layer key/value prefixes.

\subsection{Vision--Language Teacher}
Following~\cite{benavent2025text}, we use frozen CLIP encoders for frames and captions and learn lightweight temporal/projection modules to align video and text in a shared space, as illustrated in Stage~1 of Fig.~\ref{fig:framework}.

\subsubsection{Vision Branch}
For the $i$-th video, we uniformly sample $L$ frames $\{x_{i,1},\dots,x_{i,L}\}$. Each frame is encoded by the frozen CLIP image encoder $\phi_v(\cdot)$ to obtain $\ell_2$-normalized features $\mathbf{z}_{i,j}=\phi_v(x_{i,j})\in\mathbb{R}^D$. We form $\mathbf{Z}_i\in\mathbb{R}^{L\times D}$, add positional embeddings $\mathbf{P}$, and aggregate temporally using a Transformer $g_\theta(\cdot)$:
\begin{equation}
    \mathbf{H}_i = g_\theta(\mathbf{Z}_i + \mathbf{P}).
\end{equation}
Temporal average pooling yields $\mathbf{u}_i=\frac{1}{L}\sum_{j=1}^L\mathbf{H}_{i,j}$, which is projected and normalized to the final video embedding $\mathbf{v}_i\in\mathbb{R}^d$.

\subsubsection{Text Branch}
For caption $y_i$, we extract $\mathbf{q}_i=\phi_t(y_i)\in\mathbb{R}^D$ using the frozen CLIP text encoder and project it to $\mathbf{t}_i\in\mathbb{R}^d$ with $\ell_2$ normalization.

\subsubsection{Contrastive Objective}
Given a minibatch of $N$ pairs $\{(\mathbf{v}_i,\mathbf{t}_i)\}_{i=1}^N$, we compute cosine similarities $s_{ij}=\mathbf{v}_i^\top\mathbf{t}_j/\tau$ and minimize a symmetric InfoNCE loss:
\begin{equation}
    \begin{aligned}
        \mathcal{L}_{\mathrm{con}}^{v\to t}
        &= -\frac{1}{N}\sum_{i=1}^N
        \log \frac{\exp(s_{ii})}{\sum_{j=1}^N \exp(s_{ij})}, \\
        \mathcal{L}_{\mathrm{con}}^{t\to v}
        &= -\frac{1}{N}\sum_{i=1}^N
        \log \frac{\exp(s_{ii})}{\sum_{j=1}^N \exp(s_{ji})}, \\
        \mathcal{L}_{\mathrm{con}}
        &= \tfrac{1}{2}\big(
        \mathcal{L}_{\mathrm{con}}^{v\to t}
        +
        \mathcal{L}_{\mathrm{con}}^{t\to v}
        \big).
    \end{aligned}
\end{equation}

\subsubsection{Mirror-Consistency Loss}
To disambiguate mirrored semantics (e.g., left/right limbs), we create a mirrored pair $(\tilde{x}_i,\tilde{y}_i)$ by horizontally flipping frames and swapping directional words in the caption. With margin $m>0$, we enforce that each visual embedding matches its correct caption more than its mirrored caption:
\begin{equation}
    \begin{aligned}
    \mathcal{L}_{\mathrm{mc}}^{(i)}
    &=
    \max\bigl(0,\, m + s(\mathbf{v}_i,\tilde{\mathbf{t}}_i) - s(\mathbf{v}_i,\mathbf{t}_i)\bigr) \\
    &\quad +
    \max\bigl(0,\, m + s(\tilde{\mathbf{v}}_i,\mathbf{t}_i) - s(\tilde{\mathbf{v}}_i,\tilde{\mathbf{t}}_i)\bigr).
    \end{aligned}
    \label{eq:mirror_mc}
\end{equation}
The teacher objective is
$
    \mathcal{L}_{\mathrm{teacher}} = \mathcal{L}_{\mathrm{con}} + \lambda_{\mathrm{mc}}\mathcal{L}_{\mathrm{mc}}
$.

\subsection{CSI--Text Student}
The student is trained in two steps, as illustrated in Stage~2 of Fig.~\ref{fig:framework}: (1) align CSI embeddings to the frozen teacher's visual embeddings; and (2) align CSI to text with contrastive learning and mirror-consistency.

\subsubsection{Vision--CSI Alignment}
For receiver $r$, we denote CSI amplitude/phase as $\mathbf{M}_i^{(r)},\boldsymbol{\Phi}_i^{(r)}\in\mathbb{R}^{T\times N_a\times N_{\mathrm{sc}}}$. We encode them with two ResNet-18 backbones $f_{\mathrm{amp}}$ and $f_{\mathrm{pha}}$ (no weight sharing) and apply global average pooling to obtain $\mathbf{a}_i^{(r)},\mathbf{p}_i^{(r)}\in\mathbb{R}^{d_c}$:
\begin{align}
    \mathbf{a}_i^{(r)} &= f_{\mathrm{amp}}\!\left(\mathbf{M}_i^{(r)}\right), \\
    \mathbf{p}_i^{(r)} &= f_{\mathrm{pha}}\!\left(\boldsymbol{\Phi}_i^{(r)}\right).
\end{align}
We fuse amplitude and phase via a gating module~\cite{arevalo2017gmu}. Given $\mathbf{u}_i^{(r)}=[\mathbf{a}_i^{(r)};\mathbf{p}_i^{(r)}]$, a gate $\mathbf{g}_i^{(r)}\in[0,1]^{d_c}$ produces
\begin{equation}
    \mathbf{f}_i^{(r)} = \mathbf{g}_i^{(r)}\odot\mathbf{a}_i^{(r)} + (\mathbf{1}-\mathbf{g}_i^{(r)})\odot\mathbf{p}_i^{(r)}.
\end{equation}
A projection $W_p$ maps $\mathbf{f}_i^{(r)}$ to $\mathbf{c}_i^{(r)}=\mathrm{norm}(W_p\mathbf{f}_i^{(r)})\in\mathbb{R}^d$. We average valid receiver views and normalize to obtain $\bar{\mathbf{c}}_i$, then align $\bar{\mathbf{c}}_i$ to the teacher visual embedding $\mathbf{v}_i$ using a symmetric InfoNCE distillation loss $\mathcal{L}_{\mathrm{align}}$.

\subsubsection{CSI--Text Alignment}
We further align $\bar{\mathbf{c}}_i$ to the CLIP text embedding $\mathbf{t}_i$ with a symmetric InfoNCE loss $\mathcal{L}_{\mathrm{con}}$. To improve direction sensitivity, we apply a text-only mirror-consistency loss using swapped captions $\tilde{y}_i$:
\begin{equation}
\mathcal{L}_{\mathrm{mc}}^{(i)} = \max\!\bigl(0,\, m + s(\bar{\mathbf{c}}_i,\tilde{\mathbf{t}}_i) - s(\bar{\mathbf{c}}_i,\mathbf{t}_i)\bigr).
\end{equation}
The student objective is
$
    \mathcal{L}_{\mathrm{student}} = \mathcal{L}_{\mathrm{con}} + \lambda_{\mathrm{mc}}\,\mathcal{L}_{\mathrm{mc}}
$.

\begin{figure}
    \centering
    \includegraphics[width=0.4\textwidth]{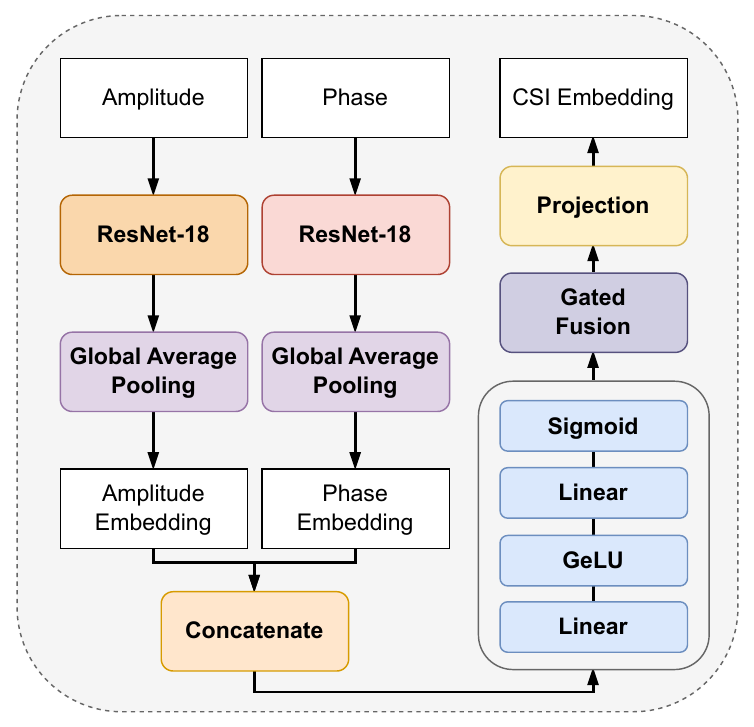}
    \caption{\textbf{CSI encoder.} Dual ResNet-18 backbones encode amplitude and phase inputs, followed by gated fusion and projection to a CSI embedding.}
    \label{fig:student}
\end{figure}

\subsection{Prefix-Guided Language Generation}
We condition a frozen GPT-2 decoder on CSI by prefix tuning. Given CSI embedding $\bar{\mathbf{c}}_i\in\mathbb{R}^d$, a lightweight MLP $g_\phi$ produces layer-wise key/value prefixes injected into each self-attention block. Let $L$ be the number of layers, prefix length $L_p$, and hidden size $d_h$ ($d_h=768$). We reshape $g_\phi(\bar{\mathbf{c}}_i)\in\mathbb{R}^{L\times L_p\times 2d_h}$ into prefixes $\{K_i^{(\ell)},V_i^{(\ell)}\}_{\ell=1}^L$.

For tokenized caption $\mathbf{y}_i=(y_{i,1},\dots,y_{i,T_i})$, the conditional likelihood is
\begin{equation}
p_\theta(\mathbf{y}_i\mid\bar{\mathbf{c}}_i) = \prod_{t=1}^{T_i} p_\theta\bigl(y_{i,t}\mid y_{i,1:t-1},\, g_\phi(\bar{\mathbf{c}}_i)\bigr),
\end{equation}
and we minimize the standard autoregressive loss
\begin{equation}
\mathcal{L}_{\mathrm{LM}} = -\frac{1}{N}\sum_{i=1}^N\sum_{t=1}^{T_i} \log p_\theta\bigl(y_{i,t}\mid y_{i,1:t-1},\, g_\phi(\bar{\mathbf{c}}_i)\bigr).
\label{eq:lm_loss}
\end{equation}
In practice, GPT-2 remains frozen and we optimize only $g_\phi$.
\section{Experiment Results}

\subsection{Main Results on the WiFi2Cap Dataset}
\label{subsec:main-results}
We first evaluate caption generation on the proposed \emph{WiFi2Cap} dataset, which is specifically constructed to support CSI-based semantic captioning. We compare two systems: (i) a baseline (CSI$\to$LM) that directly conditions the language model on the CSI encoder without vision--language alignment or mirror-consistency training; and (ii) the full WiFi2Cap pipeline.

\begin{table}[t]
    \caption{Captioning results on the WiFi2Cap and Person-in-WiFi~3D datasets. Best scores within each dataset block are in \textbf{bold}.}
    \label{tab:main-transfer-results}
    \centering
    \scriptsize
    \renewcommand{\arraystretch}{1.1}
    \resizebox{\linewidth}{!}{
    \begin{tabular}{llrrrrr}
        \toprule
        Dataset & Method & BLEU-4 & METEOR & ROUGE-L & CIDEr & SPICE \\
        \midrule
        \multirow{2}{*}{WiFi2Cap} & Baseline (CSI$\to$LM) & 14.85 & 25.86 & 32.38 & 0.12 & 0.28 \\
         & Full WiFi2Cap & \textbf{51.78} & \textbf{57.48} & \textbf{64.32} & \textbf{0.52} & \textbf{0.63} \\
        \midrule
        \multirow{2}{*}{Person-in-WiFi~3D} & Baseline (CSI$\to$LM) & 12.15 & 20.38 & 26.12 & 0.10 & 0.23 \\
         & Full WiFi2Cap & \textbf{47.07} & \textbf{58.80} & \textbf{57.26} & \textbf{0.43} & \textbf{0.51} \\
        \bottomrule
    \end{tabular}
    }
\end{table}

Table~\ref{tab:main-transfer-results} shows that WiFi2Cap clearly outperforms the baseline on all captioning metrics on the proposed \emph{WiFi2Cap} dataset. BLEU-4 improves from 14.85 to 51.78, METEOR from 25.86 to 57.48, and ROUGE-L from 32.38 to 64.32, with consistent gains in CIDEr and SPICE. These results indicate that the proposed teacher-guided alignment and prefix-based generation strategy substantially improve CSI-to-text captioning quality on our dataset.

\subsection{Transferability on the Person-in-WiFi~3D Dataset}
\label{subsec:piw3d}
We further evaluate transferability on the public \emph{Person-in-WiFi~3D} dataset~\cite{yan2024personwifi3d}. To avoid confounding factors from multi-person scenes and strong view changes, we select a single indoor scene and filter clips to a single-person subset. Each clip is associated with one of eight action categories. Because the dataset does not provide free-form captions, we convert these category labels into natural sentences using a lightweight GPT prompting recipe (1 sentence, present tense, concise, no background details) and manually spot-check the outputs. We reuse the same architectures and hyperparameters as in Sec.~\ref{subsec:main-results}.

As shown in Table~\ref{tab:main-transfer-results}, WiFi2Cap also generalizes well to this external dataset. Relative to the baseline, WiFi2Cap improves BLEU-4 from 12.15 to 47.07, METEOR from 20.38 to 58.80, and ROUGE-L from 26.12 to 57.26, with consistent improvements in CIDEr and SPICE. These results suggest that the proposed alignment-then-generation pipeline remains effective under a different capture setup, even when captions are synthesized from categorical labels.

\subsection{Qualitative Validation and Examples}
Representative qualitative examples are summarized in Table~\ref{tab:qualitative-examples}. We show one successful case and one partial mismatch case. In the successful case, the generated caption preserves the core action semantics and differs only in a near-synonymous verb choice. In the partial mismatch case, the model correctly captures the pose and balance but misses the directional phrase. Overall, these examples suggest that WiFi2Cap usually produces specific and fluent action descriptions, while remaining errors are mainly fine-grained directional confusions.

\begin{table}[t]
    \caption{Qualitative examples of generated captions.}
    \label{tab:qualitative-examples}
    \centering
    \scriptsize
    \renewcommand{\arraystretch}{1.3} % 稍微增加行高，閱讀更舒適
    \begin{tabular}{p{0.15\linewidth} p{0.35\linewidth} p{0.35\linewidth}}
        \toprule
        & \textbf{Ground Truth} & \textbf{Prediction} \\
        \midrule
        \textbf{Sample 1} \par (correct) & 
        The person raises both arms and moves them back and forth in a synchronized waving motion while staying in a squat. & 
        The person raises both arms and \textbf{swings} them back and forth in a synchronized waving motion while staying in a squat. \\
        \midrule
        \textbf{Sample 2} \par (partially correct) & 
        A man stands upright on his left foot, \textbf{facing forward}, and maintains his balance with a steady posture. & 
        A man stands upright on his left foot, \textbf{facing away}, while maintaining his balance and keeping his body steady. \\
        \bottomrule
    \end{tabular}
\end{table}

\subsection{Ablation Study}
\label{subsec:ablation}
We analyze the following axes: (i) the relative importance of different training stages; (ii) the contribution of the mirror-consistency loss; (iii) the choice of language model for prefix-guided caption generation; and (iv) the text-side backbone (CLIP B/32 vs.\ L/14). All results are reported using BLEU-4, METEOR, ROUGE-L, CIDEr, and SPICE.

\subsubsection{Ablation of Training Stages}
Table~\ref{tab:ablate-stages} quantifies the contribution of each stage. Using only Stage~2-2 (CSI--Text Alignment) performs poorly (e.g., BLEU-4 = 22.57, METEOR = 35.02, ROUGE-L = 40.39). Adding Stage~1 (Vision--Language Teacher), Stage~2-1 (Vision--CSI Alignment), and Stage~3 (Prefix-Guided Tuning) yields a large jump (BLEU-4 = 47.10, METEOR = 51.08, ROUGE-L = 57.74). Enabling the full pipeline brings the best scores (BLEU-4 = 51.78, METEOR = 57.48, ROUGE-L = 64.32), showing that the stages are complementary: the teacher establishes a stable semantic target, the student aligns CSI to text, and the prefix injects CSI semantics directly into generation.

\begin{table}[t]
    \caption{Ablation on training stages. A check mark indicates the stage is enabled.}
    \label{tab:ablate-stages}
    \centering
    \renewcommand{\arraystretch}{1.2}
    \resizebox{\linewidth}{!}{
    \begin{tabular}{ccccrrrrr}
        \toprule
        \multicolumn{4}{c}{Stages} & \multicolumn{5}{c}{Metrics} \\
        \cmidrule(lr){1-4} \cmidrule(lr){5-9}
        S1 & S2-1 & S2-2 & S3 & BLEU-4 & METEOR & ROUGE-L & CIDEr & SPICE \\
        \midrule
        --         & --         & \checkmark & \checkmark & 22.57 & 35.02 & 40.39 & 0.21 & 0.38 \\
        \checkmark & \checkmark & --         & \checkmark & 47.10 & 51.08 & 57.74 & 0.47 & 0.61 \\
        \checkmark & \checkmark & \checkmark & \checkmark & \textbf{51.78} & \textbf{57.48} & \textbf{64.32} & \textbf{0.52} & \textbf{0.63} \\
        \bottomrule
    \end{tabular}
    }
\end{table}

\subsubsection{Mirror-Consistency Analysis}
Table~\ref{tab:ablate-mirror} compares training without the mirror-consistency loss, with teacher-only mirroring (Stage~1 only), and with full mirror-consistency (Stages~1 and 2). The mirror term yields consistent improvements in overall caption quality while specifically helping preserve direction-sensitive semantics, such as left/right limb usage and mirrored action directions. By lifting all five captioning metrics from the no-mirror baseline to the full model, these results support mirror-consistency as a core component for semantic disambiguation rather than a minor auxiliary regularizer.

\begin{table}[t]
    \caption{Effect of the mirror-consistency loss.}
    \label{tab:ablate-mirror}
    \centering
    \renewcommand{\arraystretch}{1.2}
    \resizebox{\linewidth}{!}{
    \begin{tabular}{lrrrrr}
        \toprule
        Setting & BLEU-4 & METEOR & ROUGE-L & CIDEr & SPICE \\
        \midrule
        w/o Mirror-Consistency & 35.25 & 44.53 & 51.04 & 0.36 & 0.46 \\
        Teacher-only Mirror & 44.38 & 52.29 & 58.75 & 0.47 & 0.59 \\
        Full Mirror-Consistency & \textbf{51.78} & \textbf{57.48} & \textbf{64.32} & \textbf{0.52} & \textbf{0.63} \\
        \bottomrule
    \end{tabular}
    }
\end{table}

\subsubsection{Choice of Language Model}
Table~\ref{tab:ablate-lm} compares GPT-2, Qwen, and microsoft/phi-2 as the decoder in Stage~3. Qwen attains the strongest overall caption quality (highest BLEU-4 and METEOR, with solid CIDEr/SPICE), GPT-2 achieves the best ROUGE-L but lags on BLEU-4/METEOR, and phi-2 trails on most metrics. These trends indicate that both model capacity and pretraining data materially affect CSI-conditioned generation.

\begin{table}[t]
    \caption{Effect of the language model in Stage~3 (CSI-conditioned generation).}
    \label{tab:ablate-lm}
    \centering
    \renewcommand{\arraystretch}{1.2}
    \resizebox{\linewidth}{!}{
    \begin{tabular}{lrrrrr}
        \toprule
        Language Model & BLEU-4 & METEOR & ROUGE-L & CIDEr & SPICE \\
        \midrule
        GPT-2 & 51.78 & 57.48 & \textbf{64.32} & 0.52 & 0.63 \\
        Microsoft/phi-2 & 47.97 & 52.75 & 59.55 & 0.46 & 0.62 \\
        Qwen & \textbf{55.11} & \textbf{60.13} & 63.83 & \textbf{0.53} & \textbf{0.66} \\
        \bottomrule
    \end{tabular}
    }
\end{table}

\subsubsection{Text-Side Backbone}
Table~\ref{tab:ablate-textbackbone} contrasts CLIP B/32 vs.\ L/14 on the text side. L/14 gives small but consistent gains over B/32 across BLEU-4, METEOR, ROUGE-L, CIDEr, and SPICE while preserving the same ranking among language models. This suggests that a stronger text backbone modestly improves the semantic target used to supervise CSI.

\begin{table}[t]
    \caption{Choice of backbone for the text encoder.}
    \label{tab:ablate-textbackbone}
    \centering
    \renewcommand{\arraystretch}{1.2}
    \resizebox{\linewidth}{!}{
    \begin{tabular}{lccccc}
        \toprule
        Text Backbone & BLEU-4 & METEOR & ROUGE-L & CIDEr & SPICE \\
        \midrule
        CLIP B/32 & 51.78 & 57.48 & 64.32 & 0.52 & 0.63 \\
        CLIP L/14 & \textbf{53.76} & \textbf{58.37} & \textbf{65.21} & \textbf{0.53} & \textbf{0.67} \\
        \bottomrule
    \end{tabular}
    }
\end{table}

\subsection{Additional Analysis}

\subsubsection{Left--Right Confusion Analysis of VLMs}
\label{sec:left-right-confusion}
To further verify that direction-sensitive ambiguity is a real failure mode rather than an artifact of our dataset, we conduct a hand-side recognition test on Qwen2-VL-2B-Instruct. We query the model with a binary question: \emph{``Is the person using their left hand or right hand? Answer with `left' or `right'.''} Using the original pretrained weights, Qwen2-VL-2B-Instruct attains 53.3\% accuracy, indicating substantial ambiguity in distinguishing mirrored limb semantics. We then fine-tune Qwen2-VL-2B-Instruct on the UTD-MHAD dataset using the proposed Mirror-Consistency loss by pairing each training image with its horizontally flipped version. The accuracy increases to 73.3\%, demonstrating that mirror-consistency supervision effectively improves direction-sensitive understanding and supporting our use of this objective in WiFi2Cap.

\subsubsection{Visualization of Modality Alignment}
\begin{figure}[t]
  \includegraphics[width=\linewidth]{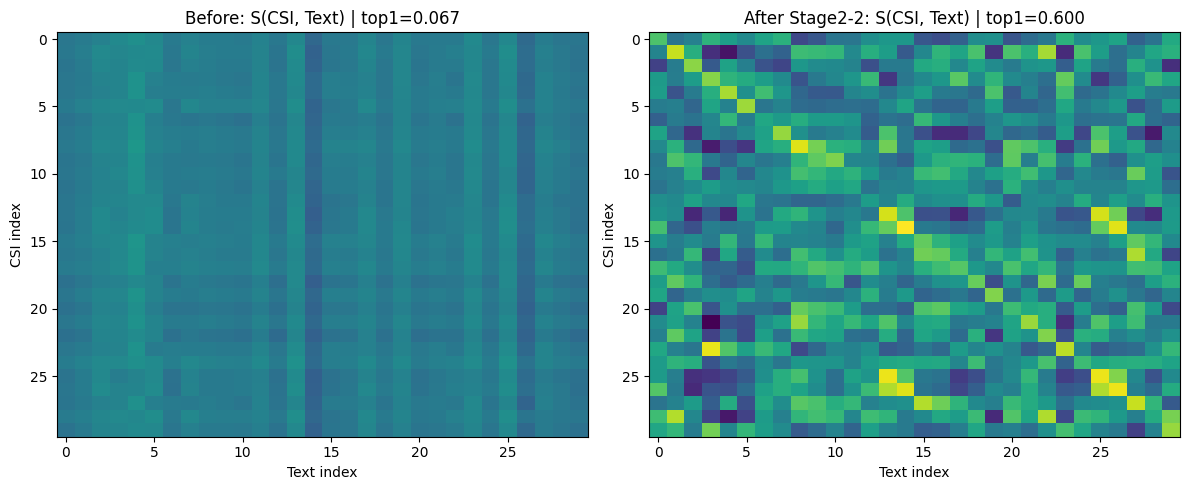}
  \caption{\textbf{Visualization of modality alignment between CSI and text embeddings.} Each row denotes a CSI embedding and each column denotes a text embedding; brighter values indicate higher cosine similarity. A clearer diagonal pattern after training indicates stronger CSI--text correspondence.}
  \label{fig:output}
\end{figure}

Fig.~\ref{fig:output} visualizes the cosine similarity matrix between CSI and text embeddings.
Before Stage~2-2, the matrix shows weak structure and a low top-1 matching accuracy of 0.067. After the proposed training, the diagonal becomes much clearer and the top-1 accuracy increases to 0.600, indicating substantially improved alignment between CSI and text representations.

\section{Conclusions}

We introduced \textbf{WiFi2Cap}, a three-stage framework for generating natural-language action descriptions directly from Wi-Fi CSI. The framework addresses two central challenges in CSI-based semantic captioning: bridging the modality gap between wireless signals and language, and preserving direction-sensitive semantics such as left/right limb descriptions. To this end, WiFi2Cap transfers semantic supervision from synchronized video--text pairs to CSI through teacher-guided alignment and CSI--text contrastive learning, while the proposed Mirror-Consistency Loss mitigates mirrored-action and left--right ambiguities during cross-modal alignment. We also introduced the \emph{WiFi2Cap Dataset}, a synchronized CSI--RGB--sentence benchmark for semantic captioning from Wi-Fi signals. Experiments and ablations show consistent gains in caption quality and direction-sensitive disambiguation. Overall, WiFi2Cap establishes a privacy-friendly bridge from wireless sensing to fine-grained semantic understanding.

\bibliographystyle{IEEEtran}
\bibliography{ref}

@inproceedings{chiu2023privacy,
  title={Privacy-preserving video conferencing via thermal-generative images},
  author={Chiu, Sheng--Yang and Huang, Yu--Ting and Lin, Chieh--Ting and Tseng, Yu--Chee and Chen, Jen--Jee and Tu, Meng--Hsuan and Tung, Bo--Chen and Nieh, YuJou},
  booktitle={2023 IEEE International Conference on Robotics and Automation (ICRA)},
  pages={9478--9485},
  year={2023},
  organization={IEEE}
}

@inproceedings{li2024mmskeleton,
  title={mmSkeleton: 3D Human Skeleton Estimation Using Millimeter Wave Radar Sparse Point Clouds},
  author={Li, Wei and Lei, Wen and Shi, Kun and Shi, Zhiguo and Wang, Yong and Zhou, Jinhai},
  booktitle={2024 IEEE/CIC International Conference on Communications in China (ICCC)},
  pages={307--312},
  year={2024},
  organization={IEEE}
}

@article{sengupta2020mm,
  title={mm-Pose: Real-time human skeletal posture estimation using mmWave radars and CNNs},
  author={Sengupta, Arindam and Jin, Feng and Zhang, Renyuan and Cao, Siyang},
  journal={IEEE sensors journal},
  volume={20},
  number={17},
  pages={10032--10044},
  year={2020},
  publisher={IEEE}
}

@inproceedings{ryoo2017privacy,
  title={Privacy-preserving human activity recognition from extreme low resolution},
  author={Ryoo, Michael and Rothrock, Brandon and Fleming, Charles and Yang, Hyun Jong},
  booktitle={Proceedings of the AAAI conference on artificial intelligence},
  volume={31},
  number={1},
  year={2017}
}

@article{gade2014thermal,
  title={Thermal cameras and applications: a survey},
  author={Gade, Rikke and Moeslund, Thomas B},
  journal={Machine vision and applications},
  volume={25},
  number={1},
  pages={245--262},
  year={2014},
  publisher={Springer}
}

@article{ren2022gopose,
  title={GoPose: 3D human pose estimation using WiFi},
  author={Ren, Yili and Wang, Zi and Wang, Yichao and Tan, Sheng and Chen, Yingying and Yang, Jie},
  journal={Proceedings of the ACM on Interactive, Mobile, Wearable and Ubiquitous Technologies},
  volume={6},
  number={2},
  pages={1--25},
  year={2022},
  publisher={ACM New York, NY, USA}
}

@inproceedings{wang2019personwifi,
  author    = {Fei Wang and Sanping Zhou and Stanislav Panev and Jinsong Han and Dong Huang},
  title     = {Person-in-WiFi: Fine-Grained Person Perception Using WiFi},
  booktitle = {Proceedings of the IEEE/CVF ICCV},
  year      = {2019},
  pages     = {5451--5460},
  doi       = {10.1109/ICCV.2019.00555}
}

@inproceedings{yan2024personwifi3d,
  author    = {Kangwei Yan and Fei Wang and Bo Qian and Han Ding and Jinsong Han and Xing Wei},
  title     = {Person-in-WiFi 3D: End-to-End Multi-Person 3D Pose Estimation with Wi-Fi},
  booktitle = {Proceedings of the IEEE/CVF Conference on CVPR},
  year      = {2024},
  pages     = {969--978},
  doi       = {10.1109/CVPR52733.2024.00098}
}

@article{zhuang2020transfer,
  author  = {Fuzhen Zhuang and Zhiyuan Qi and Keyu Duan and Dongbo Xi and Yongchun Zhu and Hengshu Zhu and Hui Xiong and Qing He},
  title   = {A Comprehensive Survey on Transfer Learning},
  journal = {Proceedings of the IEEE},
  year    = {2020},
  volume  = {PP},
  pages   = {1--34},
  doi     = {10.1109/JPROC.2020.3004555}
}

@inproceedings{radford2021learning,
  title={Learning transferable visual models from natural language supervision},
  author={Radford, Alec and Kim, Jong Wook and Hallacy, Chris and Ramesh, Aditya and Goh, Gabriel and Agarwal, Sandhini and Sastry, Girish and Askell, Amanda and Mishkin, Pamela and Clark, Jack and others},
  booktitle={International conference on machine learning},
  pages={8748--8763},
  year={2021},
  organization={PmLR}
}

@article{nguyen2024noncontact,
  author  = {Le Nguyen and Praneeth Susarla and Anirban Mukherjee and Manuel Cañellas and Constantino Álvarez Casado and Xiaoting Wu and Olli Silvén and Dinesh Jayagopi and Miguel Bordallo Lopez},
  title   = {Non-Contact Multimodal Indoor Human Monitoring Systems: A Survey},
  journal = {Information Fusion},
  year    = {2024},
  volume  = {110},
  pages   = {102457},
  doi     = {10.1016/j.inffus.2024.102457}
}

@article{biase2022markerless,
  author  = {Lazzaro Biase and Pasquale Pecoraro and Giovanni Pecoraro and Maria Letizia Caminiti and Vincenzo Di Lazzaro},
  title   = {Markerless Radio Frequency Indoor Monitoring for Telemedicine: Gait Analysis, Indoor Positioning, Fall Detection, Tremor Analysis, Vital Signs and Sleep Monitoring},
  journal = {Sensors},
  year    = {2022},
  volume  = {22},
  pages   = {8486},
  doi     = {10.3390/s22218486}
}

@inproceedings{jiang2020wifi3d,
  author    = {Wenjun Jiang and Hongfei Xue and Chenglin Miao and Shiyang Wang and Sen Lin and Chong Tian and Srinivasan Murali and Haochen Hu and Zhi Sun and Lu Su},
  title     = {Towards 3D Human Pose Construction Using WiFi},
  booktitle = {Proceedings of the 26th Annual International Conference on Mobile Computing and Networking (MobiCom)},
  year      = {2020},
  pages     = {1--14},
  doi       = {10.1145/3372224.3380900}
}

@article{hinton2015distilling,
  author  = {Geoffrey Hinton and Oriol Vinyals and Jeff Dean},
  title   = {Distilling the Knowledge in a Neural Network},
  journal = {arXiv preprint arXiv:1503.02531},
  year    = {2015},
}

@inproceedings{kim2025cosmos,
  author    = {Sanghwan Kim and Rui Xiao and Mariana-Iuliana Georgescu and Stephan Alaniz and Zeynep Akata},
  title     = {COSMOS: Cross-Modality Self-Distillation for Vision Language Pre-Training},
  booktitle = {Proceedings of the IEEE/CVF Conference on CVPR},
  year      = {2025},
  pages     = {14690--14700},
  doi       = {10.1109/CVPR52734.2025.01369}
}

@article{jiang2020picoscenes,
  author  = {Zhiping Jiang and Tom Hao Luan and Han Hao and Jing Wang and Xincheng Ren and Kun Zhao and Wei Xi and Yueshen Xu and Rui Li},
  title   = {Eliminating the Barriers: Demystify Wi-Fi Baseband Design and Introduce PicoScenes Wi-Fi Sensing Platform},
  journal = {arXiv preprint arXiv:2010.10233},
  year    = {2020},
  doi     = {10.48550/arXiv.2010.10233}
}

@article{benavent2025text,
  title={Text-driven online action detection},
  author={Benavent-Lledo, Manuel and Mulero-P{\'e}rez, David and Ortiz-Perez, David and Garcia-Rodriguez, Jose},
  journal={Integrated Computer-Aided Engineering},
  volume={32},
  number={4},
  pages={415--423},
  year={2025},
  publisher={SAGE Publications Sage UK: London, England}
}

@inproceedings{zhao2018rfpose,
  author    = {Mingmin Zhao and Tianhong Li and Mohammad Abu Alsheikh and Yonglong Tian and Hang Zhao and Antonio Torralba and Dina Katabi},
  title     = {Through-Wall Human Pose Estimation Using Radio Signals},
  booktitle = {Proceedings of the IEEE/CVF Conference on CVPR},
  year      = {2018},
  pages     = {7356--7365},
  doi       = {10.1109/CVPR.2018.00769}
}

@article{zhou2022csiformer,
  author  = {Yue Zhou and Caojie Xu and Lu Zhao and Aichun Zhu and Fangqiang Hu and Yifeng Li},
  title   = {CSI-Former: Pay More Attention to Pose Estimation with WiFi},
  journal = {Entropy},
  volume  = {25},
  number  = {1},
  pages   = {20},
  year    = {2023},
  doi     = {10.3390/e25010020}
}

@inproceedings{lester2021power,
  title={The power of scale for parameter-efficient prompt tuning},
  author={Lester, Brian and Al-Rfou, Rami and Constant, Noah},
  booktitle={Proceedings of the 2021 conference on empirical methods in natural language processing},
  pages={3045--3059},
  year={2021}
}

@inproceedings{liu2022p,
  title={P-tuning: Prompt tuning can be comparable to fine-tuning across scales and tasks},
  author={Liu, Xiao and Ji, Kaixuan and Fu, Yicheng and Tam, Weng and Du, Zhengxiao and Yang, Zhilin and Tang, Jie},
  booktitle={Proceedings of the 60th Annual Meeting of the Association for Computational Linguistics (Volume 2: Short Papers)},
  pages={61--68},
  year={2022}
}

@inproceedings{li2021prefixtuning,
  author    = {Xiang Lisa Li and Percy Liang},
  title     = {Prefix-Tuning: Optimizing Continuous Prompts for Generation},
  booktitle = {Proceedings of the 59th Annual Meeting of the Association for Computational Linguistics (ACL)},
  year      = {2021},
  pages     = {4582--4597},
  doi       = {10.18653/v1/2021.acl-long.353}
}

@article{oord2018cpc,
  author  = {Aaron van den Oord and Yazhe Li and Oriol Vinyals},
  title   = {Representation Learning with Contrastive Predictive Coding},
  journal = {arXiv preprint arXiv:1807.03748},
  year    = {2018},
  doi     = {10.48550/arXiv.1807.03748}
}

@inproceedings{chen2020simclr,
  author    = {Ting Chen and Simon Kornblith and Mohammad Norouzi and Geoffrey Hinton},
  title     = {A Simple Framework for Contrastive Learning of Visual Representations},
  booktitle = {ICML},
  year      = {2020},
}

@inproceedings{he2020moco,
  author    = {Kaiming He and Haoqi Fan and Yuxin Wu and Saining Xie and Ross Girshick},
  title     = {Momentum Contrast for Unsupervised Visual Representation Learning},
  booktitle = {Proceedings of the IEEE/CVF Conference on CVPR},
  year      = {2020},
  pages     = {9726--9735},
  doi       = {10.1109/CVPR42600.2020.00974}
}

@inproceedings{wang2015phasefi,
  author    = {Xuyu Wang and Lingjun Gao and Shiwen Mao},
  title     = {PhaseFi: Phase Fingerprinting for Indoor Localization with a Deep Learning Approach},
  booktitle = {Proceedings of the IEEE Global Communications Conference (GLOBECOM)},
  year      = {2015},
  pages     = {1--6},
  doi       = {10.1109/GLOCOM.2015.7417517}
}

@article{arevalo2017gmu,
  author  = {John Arevalo and Thamar Solorio and Manuel Montes and Fabio A. González},
  title   = {Gated Multimodal Units for Information Fusion},
  journal = {arXiv preprint arXiv:1702.01992},
  year    = {2017},
  doi     = {10.48550/arXiv.1702.01992}
}

@article{yang2024privacy,
  title={Privacy-preserving human activity sensing: A survey},
  author={Yang, Yanni and Hu, Pengfei and Shen, Jiaxing and Cheng, Haiming and An, Zhenlin and Liu, Xiulong},
  journal={High-Confidence Computing},
  volume={4},
  number={1},
  pages={100204},
  year={2024},
  publisher={Elsevier}
}

@inproceedings{pushkareva2024radar,
  title={Radar spectra-language model for automotive scene parsing},
  author={Pushkareva, Mariia and Feldman, Yuri and Domokos, Csaba and Rambach, Kilian and Di Castro, Dotan},
  booktitle={2024 International Radar Conference (RADAR)},
  pages={1--6},
  year={2024},
  organization={IEEE}
}

@article{yuan2026sig2text,
  title={Sig2text: A Vision-Language Model for Non-Cooperative Radar Signal Parsing},
  author={Yuan, Jun-Yu and Chen, Si-Yuan and Yao, Shu-Jian and Zhang, Rong and Feng, Han Cong and Jiang, Kai-li and Shang, Yu-Xin and Zhao, Yu-Xin and Tang, Bin},
  journal={IET Radar, Sonar \& Navigation},
  volume={20},
  number={1},
  pages={e70113},
  year={2026},
  publisher={Wiley Online Library}
}

\end{document}